\documentclass[10pt,twocolumn,letterpaper]{article}

\usepackage[pagenumbers]{iccv} 

\usepackage{makecell}

\definecolor{iccvblue}{rgb}{0.21,0.49,0.74}
\usepackage[pagebackref,breaklinks,colorlinks,allcolors=iccvblue]{hyperref}

\def\paperID{4916} 
\def\confName{ICCV}
\def\confYear{2025}

\title{From Gallery to Wrist: Realistic 3D Bracelet Insertion in Videos}

\author{%
    Chenjian Gao$^{1}$, Lihe Ding$^{1}$, Rui Han$^{2}$,   Zhanpeng Huang$^{2*}$, Zibin Wang$^{2}$, Tianfan Xue$^{1,3*}$ \\
  $^1$The Chinese University of Hong Kong, $^2$SenseTime Research, $^3$Shanghai AI Laboratory\\
  {\tt\small \{cjgao, dl023, tfxue\}@ie.cuhk.edu.hk} \\ 
  {\tt\small wangzb02@gmail.com, 
  \{hanrui, huangzhanpeng\}@sensetime.com, }
  \\\\\textbf{Project Page:} \url{https://cjeen.github.io/BraceletPaper/}
}

\begin{document}
\twocolumn[{%
\renewcommand\twocolumn[1][]{#1}%
\maketitle
\begin{center}
   \includegraphics[width=\linewidth]{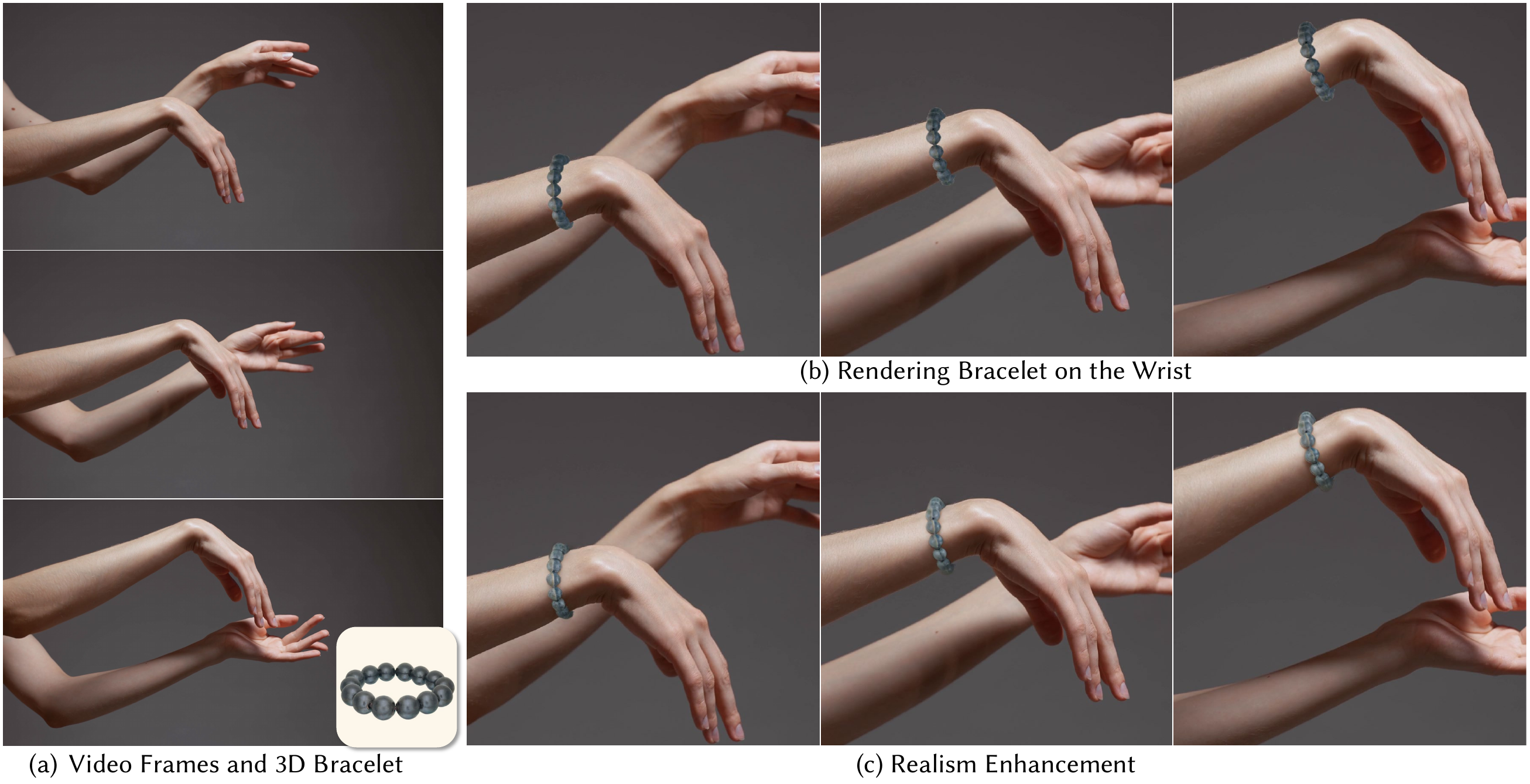}
   \captionof{figure}{
    We propose a hybrid pipeline for inserting 3D objects into videos, combining 3D Gaussian Splatting rendering for temporal consistency and a 2D diffusion-based enhancement for photorealistic lighting. In this example, a virtual bracelet is inserted onto a wrist in a dynamic scene. The 3D representation ensures temporal consistency and correct handling of occlusions as the wrist moves, while the 2D image priors enhance realism by synthesizing realistic shading. Our approach bridges the gap between 3D rendering and 2D diffusion models, achieving both temporal coherence and realism.}
   \label{fig:teaser}
\end{center}
}]
\begingroup\renewcommand\thefootnote{*}
\footnotetext{Corresponding Authors.}\endgroup
\begin{abstract}
Inserting 3D objects into videos is a longstanding challenge in computer graphics with applications in augmented reality, virtual try-on, and video composition. Achieving both temporal consistency, or realistic lighting remains difficult, particularly in dynamic scenarios with complex object motion, perspective changes, and varying illumination. While 2D diffusion models have shown promise for producing photorealistic edits, they often struggle with maintaining temporal coherence across frames. Conversely, traditional 3D rendering methods excel in spatial and temporal consistency but fall short in achieving photorealistic lighting. In this work, we propose a hybrid object insertion pipeline that combines the strengths of both paradigms. Specifically, we focus on inserting bracelets into dynamic wrist scenes, leveraging the high temporal consistency of 3D Gaussian Splatting (3DGS) for initial rendering and refining the results using a 2D diffusion-based enhancement model to ensure realistic lighting interactions. Our method introduces a shading-driven pipeline that separates intrinsic object properties (albedo, shading, reflectance) and refines both shading and sRGB images for photorealism. To maintain temporal coherence, we optimize the 3DGS model with multi-frame weighted adjustments. This is the first approach to synergize 3D rendering and 2D diffusion for video object insertion, offering a robust solution for realistic and consistent video editing.
\end{abstract}
\section{Introduction}

Inserting a 3D object into a video object is one of the classic problems in computer vision and graphics~\cite{Alex-07, Natasha-14}, and has wide applications in augmented reality, virtual try-on, and video composition. A critical challenge of this problem is to maintain both \emph{temporal consistency} and \emph{realistic lighting}. To achieve temporal consistency, the inserted object should appear coherently across frames without flickering. To achieve realistic lighting, on the other hand, the inserted object should interact naturally with the scene’s illumination and shadows, creating a convincing visual integration. While many algorithms are proposed to improve quality~\cite{chen2024anydoor, ku2024anyv2v, mou-24, ren2024consisti2v}, maintaining both temporal consistency and realistic lighting is still challenging when the object undergoes complex motions, perspective changes, or interactions with dynamic lighting and occlusions.

Recently, the rapid development of the diffusion model demonstrated the potential to increase the realism of editing video. Image diffusion models can synthesize realistic images from reference objects~\cite{chen2024anydoor, song2023objectstitch, yang-23}, and these approaches can be further extended to video object insertion by combining with an image-to-video diffusion model~\cite{ku2024anyv2v, mou-24, ren2024consisti2v, gao2025lora}. While this solution can greatly improve the visual quality, it still faces some challenges. 2D references often lack sufficient information to generate consistent novel views as the object moves through a scene. Also, when changing the object pose or camera perspective, the system must synthesize new views, which often leads to temporal incoherence, such as flickering across frames.

On the other side, traditional 3D rendering methods, commonly used in augmented reality, inherently possess the ability to maintain temporal consistency~\cite{Alex-07, Anton-07,Natasha-14}. These methods are effective in handling object motion and perspective changes. However, they often fall short in achieving photorealistic rendering due to challenges in accurately modeling complex lighting interactions. This trade-off between better temporal consistency in 3D rendering and better visual quality in 2D diffusion models suggests the need for a hybrid approach combining both paradigms.

Therefore, in this work, we propose an object insertion pipeline that utilizes both 2D diffusion models and 3D rendering techniques. In particular, we focus on the task of inserting bracelets onto a wrist in dynamic scenes. Bracelets contain complex and dynamic highlights, making it a perfect case to study how to maintain temporal consistency and realistic lighting. To combine both 2D diffusion models and 3D rendering, our approach first represents a virtual object using 3D Gaussian Splatting (3DGS)~\cite{kerbl3Dgaussians}, and generates initial inserted videos using rendered appearance. Since this initial result is based on 3D rendering, it guarantees temporal consistency. Then, to improve the realism, we further introduce a 2D diffusion-based enhancement model to refine the rendered image, ensuring that the inserted object naturally aligns with the scene’s illumination.  

One technical challenge is how the enhancement model imposes realistic lighting on the initial editing results, while still maintaining temporal consistency in the initial videos. To achieve this, we propose a shading-driven pipeline that begins with image decomposition to separate the object's appearance into intrinsic components including albedo, shading, and reflectance residual. The pipeline then operates in two stages: first, we manipulate lighting interactions in the shading domain, ensuring realistic adaptation to the scene’s illumination. Second, we refine the results in sRGB space to enhance fine details and ensure visual coherence. Our network is fine-tuned from Stable Diffusion~\cite{Rombach_2022_CVPR}, leveraging its powerful image prior. To maintain high fidelity to the input, we adopt a single-step diffusion formulation and an enhanced decoder. Finally, we leverage the 3DGS model as an intermediary to enhance temporal consistency by optimizing its color with weighted multiple frames, ensuring each frame is more influenced by temporally or visually adjacent frames for robust temporal coherence.

Experimental results show that our method achieves high realism and temporal consistency in video object insertion. Additionally, the framework’s high editability is supported by an interactive workflow, allowing users to intuitively adjust the object’s pose and placement for flexible insertion. 
\section{Related Work}
\begin{figure*}
  \centering
  \includegraphics[width=\linewidth]{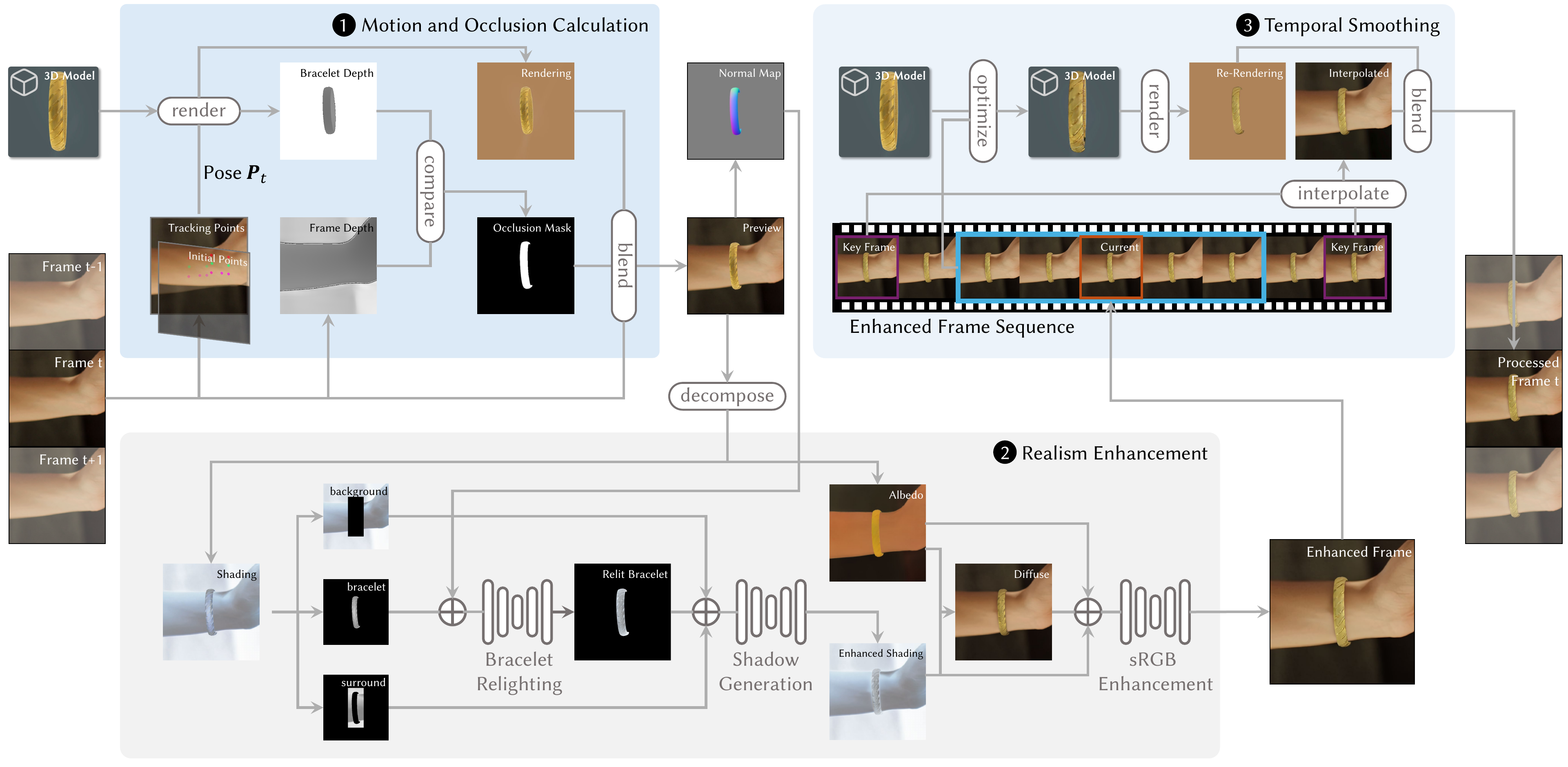}
  \caption{
    Method overview. Our pipeline inserts a 3D bracelet into a video while maintaining temporal consistency and realistic lighting. 1) We first compute motion and occlusion using 3D Gaussian Splatting (3DGS), leveraging 2D tracking points to align the bracelet with the wrist’s motion and monocular depth maps to handle occlusions. 2) Next, we enhance realism through a shading-driven approach, decomposing the image into albedo and shading components. The shading is refined using a diffusion-based model to adapt the bracelet’s lighting to the scene, while the albedo ensures color consistency. 3) Finally, we apply temporal smoothing to the bracelet and shadows, optimizing the 3DGS model and interpolating frames to ensure smooth transitions across the video.
  }
  \label{fig:pipeline_main}
\end{figure*}
\subsection{Object Insertion}
Video object insertion blends a target object into video frames with consistent motion and illumination. Early methods \cite{Alex-07, Anton-07,Natasha-14} used image warping or 3D models as geometric proxies, but their time-consuming per-video optimization limited widespread use. Recently, many works \cite{yang-23, chen2024anydoor, Winter-24, song2024imprint, zhang2023controlcom} leverage diffusion models for high-quality visual results, but often have issues with preserving the identity of the inserted object. Some works \cite{ouyang-24,mou-24, tu-25} extend object insertion to video editing, adding the challenge of motion consistency. These methods often suffer from identity and motion inconsistencies due to limited structural information from a single image. Our method, on the other hand, combines 3D rendering and 2D diffusion models, ensuring both temporal consistency and photorealistic lighting while addressing the limitations of existing approaches.

\subsection{Image harmonization}
Image Harmonization aims to adjust foreground colors and lighting to match the background, creating natural compositions. Most methods treat this as an image-to-image translation problem \cite{cong2022high, ke2022harmonizer}, trained on datasets with foreground degradation. Recent work has explored illumination-aware harmonization \cite{liao2019approximate, guo2021intrinsic, guo2022transformer, hu2021neursf, bao2022deep, bhattad2022cut, wang2023semi, careagaCompositing}. However, these methods heavily rely on geometric information to model lighting, limiting their flexibility. In contrast, our method only uses geometry to synthesize diverse lighting conditions during training, avoiding reliance on explicit geometric priors during inference. We leverage the strong priors of image diffusion models~\cite{Rombach_2022_CVPR} to perform harmonization in both the shading and RGB domains, enabling realistic adaptation of local lighting while maintaining global consistency.

\section{Methodology}

Our method leverages 3D Gaussian Splatting (3DGS) to ensure consistent object representation across all viewpoints, overcoming limitations of 2D-based approaches. The pipeline comprises three key modules: \emph{Motion and Occlusion-Aware Calculation} (Sec.~\ref{sec:motion_occlusion}), \emph{Realism Enhancement} (Sec.~\ref{sec:realism_enhance}), and \emph{Temporal Smoothing} (Sec.~\ref{sec:temporal_smooth}). As shown in Fig.~\ref{fig:pipeline_main}, the process begins by aligning the 3D bracelet with the wrist's motion and computing occlusion masks to preserve depth ordering. The bracelet is then rendered and blended into the video. Next, realism is enhanced through lighting and shadow effects using novel 2D models. Finally, temporal smoothing ensures stability across frames. This 3D-based approach avoids view synthesis artifacts, achieving high fidelity and temporal consistency in dynamic scenes. The following sections detail each module.

\subsection{Motion and Occlusion Aware Insertion}
\label{sec:motion_occlusion}
\begin{figure}[t]
  \centering
  \includegraphics[width=\linewidth]{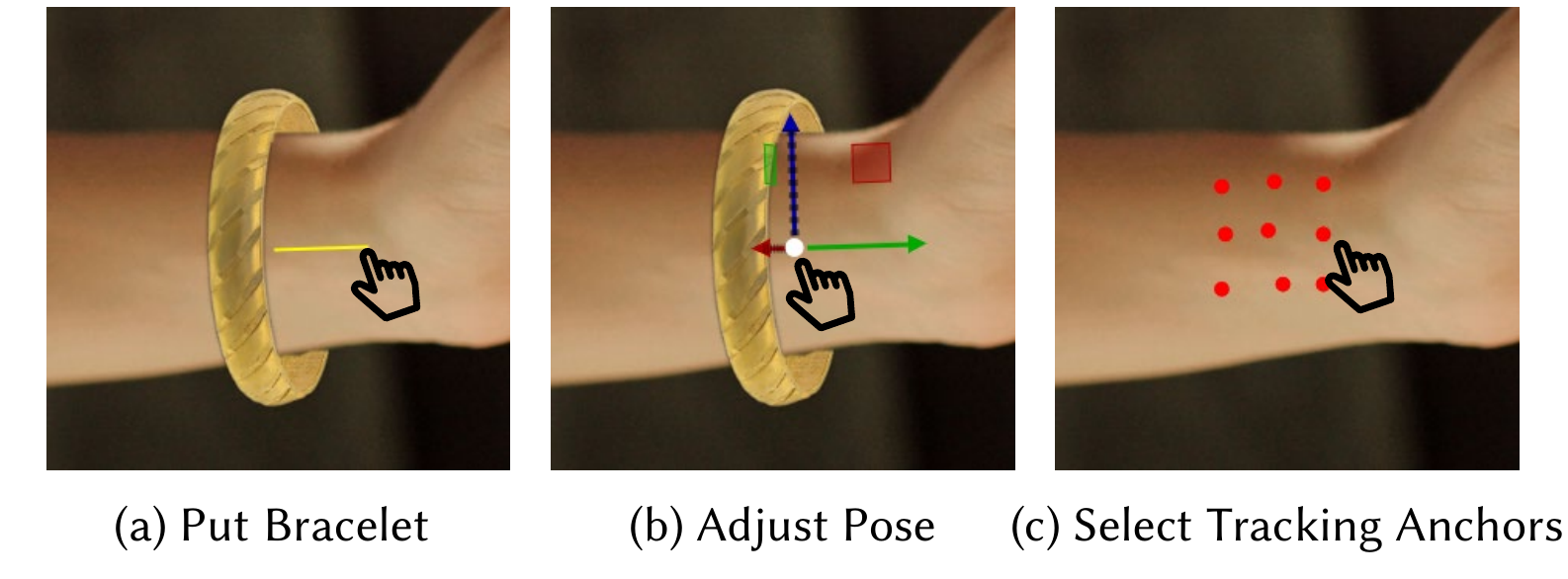}
  \caption{
    Interactive GUI. (a) Put Bracelet: Place the bracelet on the wrist and adjust its orientation. (b) Adjust Pose (optional): Fine-tune the bracelet’s position, orientation, and scale in 3D space. (c) Select Tracking Anchors (optional):  Adjust tracking anchor points to ensure robust motion tracking. 
  }
  \label{fig:gui}
\end{figure}

Our method follows a traditional AR pipeline to insert a 3D bracelet into the input video. We first render the bracelet onto the video by estimating its initial pose, which is interactively adjusted by users through a lightweight GUI (Fig. \ref{fig:gui}(a) and (b)). To track the bracelet's motion, we utilize 2D keypoint tracking on the skin via \emph{CoTracker}~\cite{karaev24cotracker3}, enabling 3D pose estimation across frames. The bracelet is then rendered as an overlay on each frame using its computed 3D poses. To handle occlusions, we generate an occlusion map per frame by comparing the bracelet's depth with the scene depth, leveraging depth estimates from \emph{UniDepth}~\cite{piccinelli2024unidepth}. Finally, we blend the rendering of the bracelet with the scene's background to obtain an initial preview. Further implementation details are provided in the supplementary material.

\subsection{Realism Enhancement}
\label{sec:realism_enhance}
The initial preview rendering generated in Sec.~\ref{sec:motion_occlusion} can correctly track object motion, but lighting is not realistic due to missing shadows, fake highlights, and inconsistencies between the original videos and the inserted bracelets. This discrepancy results in an unnatural appearance, as the bracelet fails to interact convincingly with the scene lighting, as shown Fig.~\ref{fig:teaser}. Directly modifying the lighting parameters of the 3D Gaussian Splatting of the bracelet is challenging, as the real-world lighting is hard to model.

Instead, in this work, we propose a novel 2D realism enhancement to solve this problem in 2D image space, utilizing the powerful image prior in diffusion models. The enhancement consists of \emph{bracelet relighting} and \emph{shadow generation}, ensuring the bracelet seamlessly integrates into the scene. Below, we will discuss the overview of our enhancement model (Sec.~\ref{sec:shading-driven}), network architecture (Sec.~\ref{sec:network}), training dataset (Sec.~\ref{sec:data}), and optimization target (Sec.~\ref{sec:optimization}) in details.

\subsubsection{Shading-Driven Realism Enhancement}
\label{sec:shading-driven}
A naive way to enhance lighting is to directly train an enhancement that takes an initial RGB frame as input and outputs an enhanced RGB image. However, we found that this simple sRGB enhancement does not perform well, as lighting and color information are inherently entangled, potentially leading to unrealistic color shifts or artifacts. Therefore, we proposed a novel shading-driven enhancement. It decomposes the input into shading and albedo components and only enhances shading in the first stage, which greatly reduces the risk of introducing artifacts or modifying the texture of the original 3D objects. With the initial enhancement result, we will further refine it in sRGB space.

To achieve this, given an input image \( \mathbf{I}_t \) in sRGB space, we first undo the gamma correction with an exponent of \( 2.2 \) to transform \( \mathbf{I}_t \) into linear RGB space, denoted as \(\tilde{\mathbf{I}}_t\). The linear RGB image \(\tilde{\mathbf{I}}_t\) is then decomposed into its intrinsic components using a pretrained model~\cite{careaga2024colorful}:
\[
\tilde{\mathbf{I}}_t = \mathbf{A}_t \cdot \mathbf{S}_t + \mathbf{R}_t.
\]
where \( \mathbf{A}_t \) represents the albedo (reflectance properties), \( \mathbf{S}_t \) is the shading (illumination effects), and \( \mathbf{R}_t \) captures residual details such as specular highlights. Here, \( \mathbf{A}_t \cdot \mathbf{S}_t \) corresponds to the diffuse component of the image, while \( \mathbf{R}_t \) accounts for non-lambertian lighting components.

To ensures that enhancement model does change the content of the original input video, we partition the shading map into three regions: the \emph{bracelet region} $\mathbf{M}$, which contains the bracelet itself; the \emph{background region} $\mathbf{M}_\text{bg}$, representing pixels outside the expanded bounding box of $\mathbf{M}$ to provide scene lighting context; and the \emph{surrounding region} $\mathbf{M}_\text{surr} = \mathbf{1} - \mathbf{M}_\text{bg} - \mathbf{M}$, which captures shadows cast by the bracelet. The bracelet mask $\mathbf{M}$ is obtained through 3D model segmentation, while $\mathbf{M}_\text{bg}$ is defined as the area outside the expanded bounding box of $\mathbf{M}$. The surrounding mask $\mathbf{M}_\text{surr}$ is derived as the complement of $\mathbf{M}$ and $\mathbf{M}_\text{bg}$. By multiplying the original shading map with each region mask, we obtain three distinct shading maps: 
\[
\mathbf{S}_\text{bracelet} = \mathbf{M} \cdot \mathbf{S}, \quad \mathbf{S}_\text{bg} = \mathbf{M}_\text{bg} \cdot \mathbf{S}, \quad \text{and} \quad \mathbf{S}_\text{surr} = \mathbf{M}_\text{surr} \cdot \mathbf{S}.
\]
Our enhancement model will not change the background shading $\mathbf{S}_\text{bg}$, and will treat the bracelet shading $\mathbf{S}_\text{bracelet}$ and surrounding shadow shading $\mathbf{S}_\text{surr}$ separately.

For the bracelet shading, we train a \emph{bracelet relighting network} $f_\text{br}$ that takes the bracelet shading $\mathbf{S}_\text{bracelet}$, background shading $\mathbf{S}_\text{bg}$, and surface normal map $\mathbf{N}_t$ predicted with \cite{martingarcia2024diffusione2eft} as inputs to produce the relit bracelet shading:
\[
\mathbf{S}_\text{relit} = f_\text{br}(\mathbf{S}_\text{bracelet}, \mathbf{S}_\text{bg}, \mathbf{N}_t).
\]

For the surrounding shadow shading, we train a \emph{shadow generation network} $f_\text{sh}$ that takes the relit bracelet shading $\mathbf{S}_\text{relit}$, background shading $\mathbf{S}_\text{bg}$, and surrounding shading $\mathbf{S}_\text{surr}$ as inputs, and outputs the full shading with shadows:
\[
\mathbf{S}_\text{enhanced} = f_\text{sh}(\mathbf{S}_\text{relit}, \mathbf{S}_\text{bg}, \mathbf{S}_\text{surr}).
\]

The enhanced shading $\mathbf{S}_\text{enhanced}$ is then combined with the albedo $\mathbf{A}_t$ to produce the diffuse image:
\[
\tilde{\mathbf{I}}_\text{diffuse} = \mathbf{A}_t \cdot \mathbf{S}_\text{enhanced}.
\]

Finally, the diffuse image $\tilde{\mathbf{I}}_\text{diffuse}$ is transformed back into sRGB space using gamma correction. The resulting sRGB diffuse image $\mathbf{I}_\text{diffuse}$ is refined using the \emph{sRGB enhancement network} $f_\text{sRGB}$, which enhances fine details:
\[
\mathbf{I}_\text{refined} = f_\text{sRGB}(\mathbf{I}_\text{diffuse}, \mathbf{A}_t, \mathbf{S}_\text{enhanced}).
\]

It is worth noting that recent work \cite{careagaCompositing} demonstrates the effectiveness of harmonizing in the shading domain. However, \cite{careagaCompositing} inherently suffers from limitations: 1) it cannot handle shadows cast by objects, and 2) their approximation of environmental lighting as a single light source during inference restricts their generalizability. In contrast, our method enhances realism separately in the shading domain and sRGB domain, processes shadows independently, and leverages the prior of Stable Diffusion~\cite{Rombach_2022_CVPR} (to be introduced in Sec.~\ref{sec:network}), thereby overcoming these limitations.

\subsubsection{Network Architecture}
\label{sec:network}
\begin{figure}
  \centering
  \includegraphics[width=0.9\linewidth]{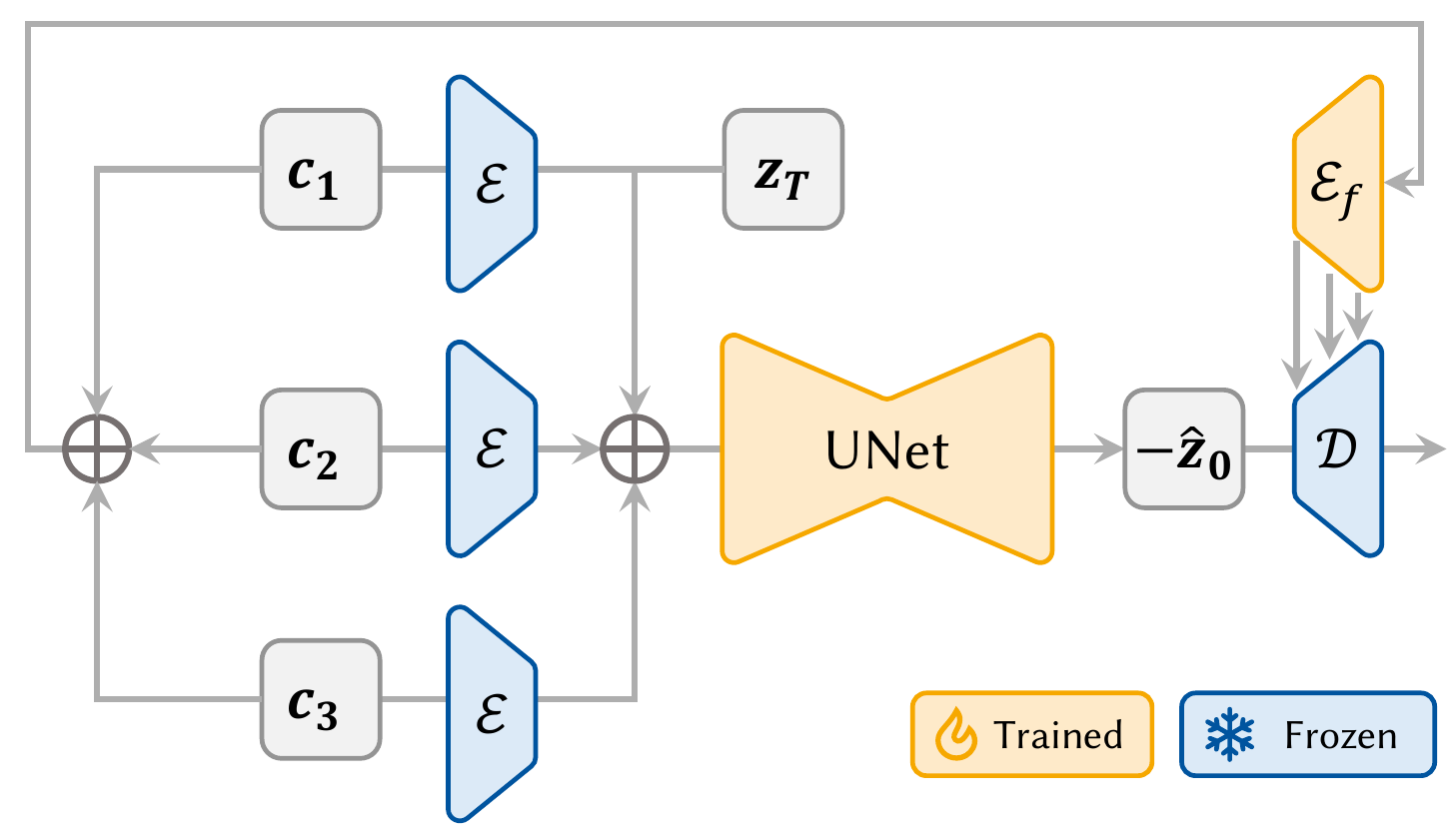}
  \caption{
    Network architecture. We adopt single-step diffusion formulation with an enhanced decoder. The UNet predicts a denoised latent in one forward pass, while multi-scale features from an additional encoder are connected to the VAE decoder.
  }
  \label{fig:network}
\end{figure}
With the shading-driven realism enhancement pipeline defined, we now describe the implementation of the neural networks \( f_\text{br} \), \( f_\text{sh} \), and \( f_\text{sRGB} \). Each network is designed to accept three input conditions denoted as \( \mathbf{c}_1, \mathbf{c}_2, \mathbf{c}_3 \) and produce a single output image. To achieve realistic lighting, we finetune \emph{Stable Diffusion} model~\cite{Rombach_2022_CVPR}, leveraging its powerful images prior. To maintain the bracelet's appearance consistent with the original rendering, we adopt single-step diffusion with the enhanced decoder as shown in Fig.~\ref{fig:network}.

\emph{Single-step Diffusion.} 
We adopt a single-step diffusion formulation~\cite{martingarcia2024diffusione2eft} to perform deterministic sampling to maintain fidelity to the input bracelet. Instead of gradually denoising over multiple steps, we concatenate a zero latent \(\mathbf{z}_T = \mathbf{0}\) with the encoded condition and feed it into the UNet~\cite{ronneberger2015u}, which is trained to predict the clean output latent in a single forward pass. The process is defined as:
\[
\hat{\mathbf{z}}_0 = - \epsilon_\theta(\mathbf{z}_T \oplus \mathcal{E}(\mathbf{c}_1)\oplus \mathcal{E}(\mathbf{c}_2)\oplus \mathcal{E}(\mathbf{c}_3)),
\]
where \( \mathcal{E}(\cdot) \) is the encoded condition, \( \epsilon_\theta \) is the UNet, \( \hat{\mathbf{z}}_0 \) is the predicted clean latent, and \(\oplus\) denotes concatenation. 

\emph{Enhanced Decoder.} The decoding process from latent space to the final result image can introduce unintended artifacts, as the VAE (Variational AutoEncoder) decoder~\cite{doersch2016tutorial} may add content that deviates from the input condition. Similar to \cite{weng2024cad}, we introduce additional control over the decoder's intermediate features by incorporating another encoder \( \mathcal{E}_f \). This encoder extracts multi-scale features \( \{\mathbf{f}^i\}_{i=1}^L \) from the input condition, where \( \mathbf{f}^i = \mathcal{E}_f^i(\mathbf{c}_1 \oplus \mathbf{c}_2 \oplus \mathbf{c}_3) \) represents the features at the \( i \)-th scale. These features are then added into the corresponding layers of the VAE decoder \( \mathcal{D} \). The decoding process can be formulated as:
\[
\mathbf{I}_\text{out} = \mathcal{D}(\hat{\mathbf{z}}_0, \{\mathbf{f}^i\}_{i=1}^L),
\]
where \( \mathcal{D} \) is the VAE decoder, \( \hat{\mathbf{z}}_0 \) is the predicted clean latent, and \( \{\mathbf{f}^i\}_{i=1}^L \) are the multi-scale features extracted by \( \mathcal{E}_f \).

\subsubsection{Training Data Generation.}
\label{sec:data}

\begin{figure}
  \centering
  \includegraphics[width=\linewidth]{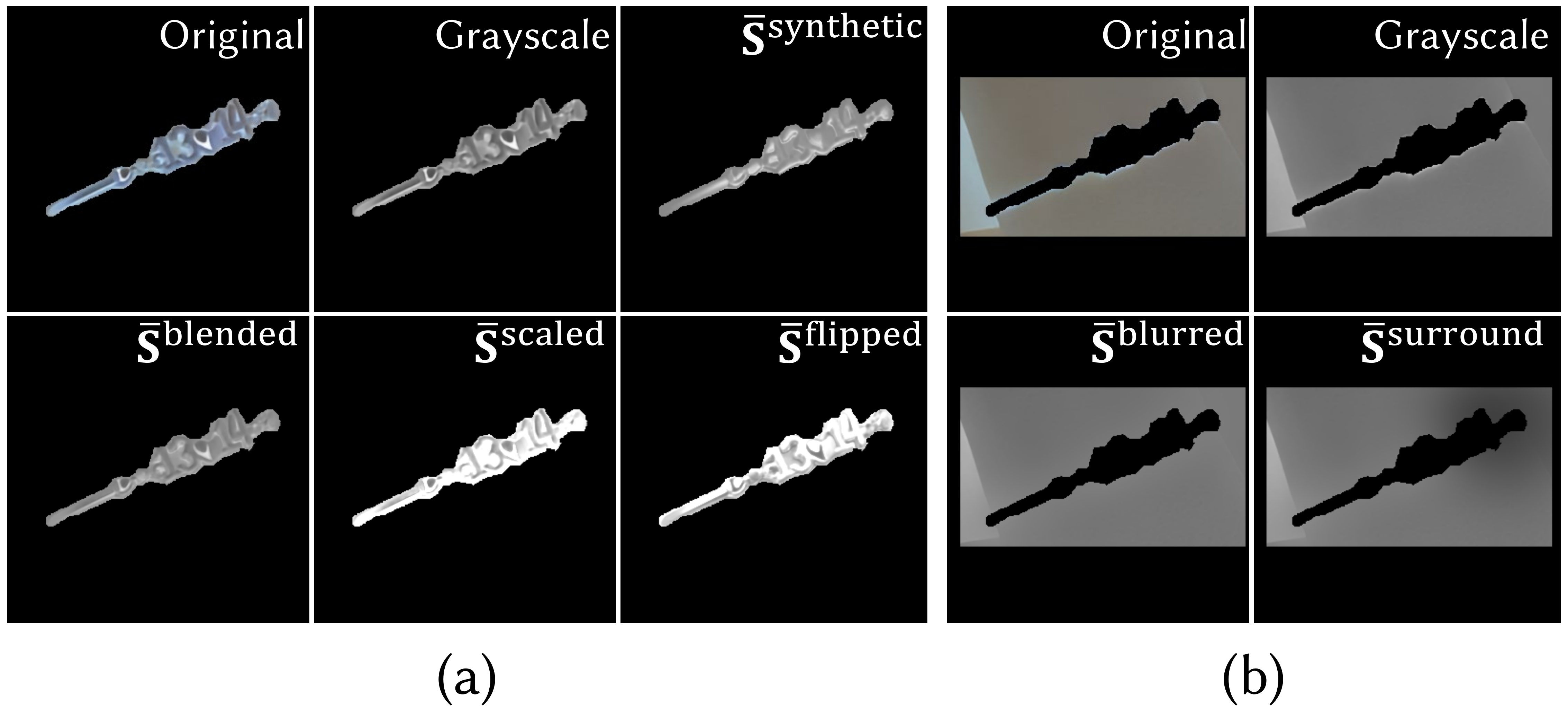}
  \caption{
    Shading maps in data augmentation. (a) Bracelet shading: Starting from the original shading, we convert it to grayscale, blend with synthetic shading, scale intensity, and flip brightness. (b) Shadow shading: We apply Gaussian blur to the grayscale shading and add random patch to surrounding region.
  }
  \label{fig:data}
\end{figure}

Collecting training data for lighting enhancement is not easy, as we do not have the pair of images before and after enhancement. Therefore, we synthesize training pairs by applying degradation (augmentation) to well-lit images and use them as the training input. To do so, we collect about \(13,000\) images featuring bracelets worn on wrists. From these images, we decompose the images into shading \( \mathbf{S}^\text{target} \) and albedo \( \mathbf{A} \). 
Then we perturb \( \mathbf{S}^\text{target} \) to construct diverse input shading maps for training. Utilizing this data, we propose two novel data simulation pipelines, for the bracelet relighting network and shadow network separately.
Fig.~\ref{fig:data} demonstrates intermediate results of the data augmentation.

\emph{Augmentation for Bracelet Relighting Network $f_\text{br}$.} To get bracelet shading under varying lighting conditions, we begin by synthesizing shading maps using the surface normal map \( \mathbf{N} \) and multiple randomly sampled light directions \( \{\mathbf{L}^\text{rand}_i\}_{i=1}^M \). For each \(\mathbf{L}^\text{rand}_i\), \( \bar{\mathbf{S}}^\text{synthetic}_i \) is computed as:
\[
\bar{\mathbf{S}}^\text{synthetic}_i = (\mathbf{N} \cdot \mathbf{L}^\text{rand}_i)^{\alpha},
\]
where \( \alpha \) is sampled from \( [1.0, 6.0] \) to control the sharpness of the shading variations. Then \( \bar{\mathbf{S}}^\text{synthetic} \) is obtained by blending the individual shading maps:
\[
\bar{\mathbf{S}}^\text{synthetic} = \sum_{i=1}^M w_i \cdot \bar{\mathbf{S}}^\text{synthetic}_i,
\]
To simulate dominant light sources, we sparsify blending weights $\{w_i\}_{i=1}^M$ using softmax with temperature $\tau$:
\[
w_i = \frac{\exp(z_i / \tau)}{\sum_{j=1}^M \exp(z_j / \tau)}.
\]
where $\{z_i\}$ are uniformly sampled.

While synthetic shading maps provide diverse lighting conditions, they lack fine material details due to the simplicity of the shading synthesis process. 
To preserve material information, we convert the original shading map \( \mathbf{S}^\text{target} \) into grayscale, denoted as \( \bar{\mathbf{S}} \), and blend it with \(\bar{\mathbf{S}}^\text{synthetic}\) using a blending weight \( \beta \) sampled uniformly from \( [0, 1] \):
\[
\bar{\mathbf{S}}^\text{blended} = \beta \cdot \bar{\mathbf{S}} + (1 - \beta) \cdot \bar{\mathbf{S}}^\text{synthetic}.
\]

To further enhance shading diversity, we scale the grayscale shading map \( \bar{\mathbf{S}}^\text{blended} \) by adjusting intensity range:
\[
\bar{\mathbf{S}}^\text{scaled} = (\bar{\mathbf{S}}^\text{blended} - m) \cdot \gamma + m.
\]
where \( m = \left(\bar{S}^\text{blended}_\text{max} + \bar{S}^\text{blended}_\text{min}\right)/2 \), and \( \gamma \) is the random scaling factor sampled uniformly from \( [0.5, 1.5] \). 

Finally, we randomly flip the brightness relationships in \( \bar{\mathbf{S}}^\text{scaled} \) with a probability of 0.5. \( \mathbf{S}^\text{flipped} \) is computed as:
\[
\mathbf{S}^\text{flipped} = \bar{S}^\text{scaled}_\text{max} - \bar{\mathbf{S}}^\text{scaled} + \bar{S}^\text{scaled}_\text{min}.
\]

\emph{Augmentation for Shadow Generation Network $f_\text{sh}$.} Since \( f_\text{sh} \) learns to add realistic shadows to skin surrounded bracelet, we also augment the shading map by perturbing shadow information near the bracelet. 

First, we apply Gaussian blur to the shading map \( \bar{\mathbf{S}} \) near the bracelet mask \( \mathbf{M} \). The blur is applied within a proximity radius \( r \) around the mask, ensuring a smooth transition between the modified and unmodified regions. The blurred shading map \( \bar{\mathbf{S}}^\text{blurred} \) is computed as:
\[
\bar{\mathbf{S}}^\text{blurred}(\mathbf{x}) = 
\begin{cases}
G_{\sigma_b} * \bar{\mathbf{S}}(\mathbf{x}), & \text{if } \|\mathbf{x} - \mathbf{M}\| \leq r, \\
\bar{\mathbf{S}}(\mathbf{x}), & \text{otherwise}.
\end{cases}
\]
where \( G_{\sigma_b} \) is a Gaussian kernel with standard deviation \( \sigma_b \).

Next, we add a randomly generated patch to the shading map. The intensity of this patch is defined by blended Gaussian kernels,  which are centered at random locations \( \{\mathbf{c}_k\}_{k=1}^K \) within the bracelet mask \( \mathbf{M} \) and its surrounding region. Each Gaussian image \( G_k \) is defined as:
\[
G_k(\mathbf{x}) = A_k \cdot \exp\left(-\frac{\|\mathbf{x} - \mathbf{c}_k\|^2}{2\sigma_k^2}\right),
\]
where \( A_k \) is the amplitude, \( \mathbf{c}_k \) is the center, and \( \sigma_k \) is the standard deviation of the \( k \)-th Gaussian. The amplitudes \( A_k \) and standard deviations \( \sigma_k \) are sampled from predefined ranges \( [A_\text{min}, A_\text{max}] \) and \( [\sigma_\text{min}, \sigma_\text{max}] \), respectively. The modified shading map \( \bar{\mathbf{S}}^\text{surround} \) is then computed as:
\[
\bar{\mathbf{S}}^\text{surround}(\mathbf{x}) = \bar{\mathbf{S}}^\text{blurred}(\mathbf{x}) + \sum_{k=1}^K G_k(\mathbf{x}).
\]

\subsubsection{Optimization Target}
\label{sec:optimization}
We train the bracelet relighting network \( f_\text{br} \), the shadow generation network \( f_\text{sh} \), and the sRGB enhancement network \( f_\text{sRGB} \) sequentially. All networks are initialized using pre-trained Stable Diffusion~\cite{Rombach_2022_CVPR} parameters, with both the UNet \( \epsilon_\theta \) and the additional encoder \(\mathcal{E}_f \) fine-tuned during training. We optimize the networks using the AdamW~\cite{loshchilov2017decoupled} optimizer with a learning rate of \( 3 \times 10^{-5} \) and a batch size of 16. The shading enhancement networks are trained with a combination of L1 loss and multiscale gradient loss~\cite{li2018megadepth}:
\[
\mathcal{L}_\text{shading} = \lambda_1 \|\mathbf{S}_t^\text{pred} - \mathbf{S}_t^\text{target}\|_1 + \lambda_2 \sum_{k=1}^K \|\nabla_k \mathbf{S}_t^\text{pred} - \nabla_k \mathbf{S}_t^\text{target}\|_1,
\]
where \( \mathbf{S}_t^\text{pred} \) is the predicted shading, \( \mathbf{S}_t^\text{target} \) is the target shading, \( \nabla_k \) denotes the gradient at scale \( k \). The sRGB enhancement network is trained using L1 loss:
\[
\mathcal{L}_\text{sRGB} = \|\mathbf{I}_t^\text{pred} - \mathbf{I}_t^\text{target}\|_1.
\]

\subsection{Temporal Smoothing}
\label{sec:temporal_smooth}
The realism enhancement introduced above only processes each frame independently, which may introduce temporal inconsistencies. To address this, we apply temporal smoothing strategies for both the bracelet and shadows, ensuring smooth transitions and coherence over time.

\emph{Bracelet Smoothing.} We optimize the 3DGS model \( \mathcal{G} \) in a frame-by-frame manner, using a sliding window to ensure smooth transitions and coherence over time. At each frame \( t \), we optimize \( \mathcal{G} \) by solving the following optimization:
\[
\mathcal{G}_t^* = \operatorname*{argmin}_{\mathcal{G}} \sum_{k=t-W/2}^{t+W/2} w(k - t) \cdot \| \mathcal{R}(\mathbf{K}, \mathbf{P}_k, \mathcal{G}) - \mathbf{I}_t^\text{refined} \|^2,
\]
where \( \mathcal{R}(\mathbf{K}, \mathbf{P}_t, \mathcal{G}) \) denotes the rendering of the 3DGS model \( \mathcal{G} \) at pose \( \mathbf{P}_t \) with camera intrinsics \( \mathbf{K} \). The window \( w(\cdot) \) is a Gaussian function centered at the current frame \( t \), assigning higher weights to frames closer to \( t \). Note that we only optimize the color attributes of the 3DGS model to keep geometric structure unchanged and preserve identity. Details on optimizing 3DGS color attributes are provided in the supplementary material. The smoothed bracelet appearance is denoted as \( \mathbf{I}_t^\text{rerender} = \mathcal{R}(\mathbf{K}, \mathbf{P}_t, \mathcal{G}_t^*) \).

\emph{Shadow Smoothing.} We use a frame interpolation method for smoothing shadows. We adopt the method from \cite{yang2023rerender}, uniformly selecting key frames from the \( \mathbf{I}_t^\text{refined} \) sequence and using \emph{EbSynth}~\cite{Jamriska2018} to interpolate the remaining frames. The final interpolated sequence is denoted as \( \mathbf{I}_t^\text{interp} \). For details of the interpolation method, please refer to \cite{yang2023rerender}.

Finally, we use the bracelet mask \( \mathbf{M}_t \) to blend the re-rendered bracelet \( \mathbf{I}_t^\text{rerender} \) and \( \mathbf{I}_t^\text{interp} \) as follows:
\[
\mathbf{I}_t^\text{smooth} = \mathbf{M} \cdot \mathbf{I}_t^\text{rerender} + (1 - \mathbf{M}) \cdot \mathbf{I}_t^\text{interp}.
\]

\begin{figure*}[t]
    \centering
    \includegraphics[width=\linewidth]{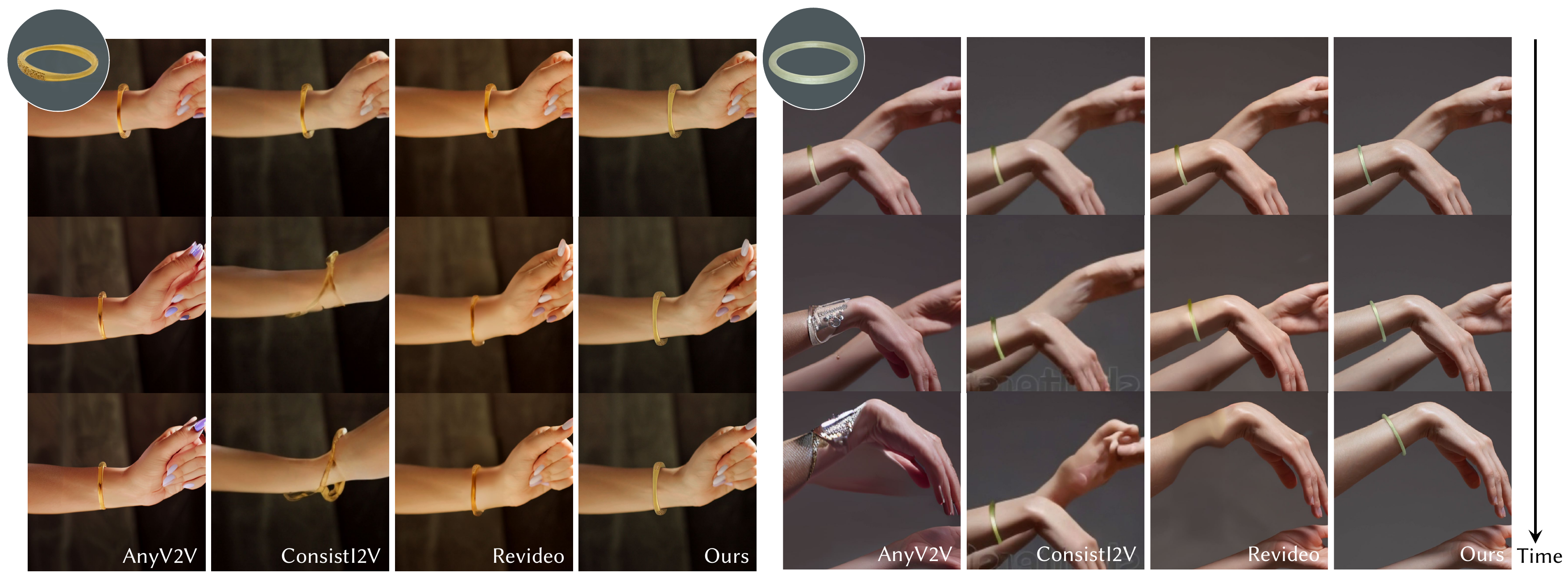}
    \caption{Qualitative comparison results. We compare our method against AnyV2V\cite{ku2024anyv2v}, ConsistI2V\cite{ren2024consisti2v} and ReVideo\cite{mou-24}. Our approach achieves superior realism, temporal consistency, and fidelity. Please zoom in for better view.}
    \label{fig:visual_comparison}
\end{figure*}
\section{Experimental Evaluation}
To evaluate the effectiveness of our proposed framework, we conduct a comprehensive set of experiments focusing on \emph{realism}, \emph{temporal consistency}, and \emph{fidelity to the input}.

\noindent\textbf{Evaluation Data. }
For evaluation, we create a dataset consisting of 3D object reconstructions and real-world video sequences. First, we capture and reconstruct 8 unique bracelets using a smartphone camera. These bracelets vary in material, texture, and geometric complexity.
Next, we select 7 video sequences from the Pexels\footnote{https://www.pexels.com}, each featuring a wrist in different environments and under varying lighting conditions. These scenes include indoor settings with artificial lighting, outdoor environments with natural sunlight, and dynamic scenarios where both the wrist and camera are in motion.
By pairing each of the 8 bracelets with the 7 videos, we create 56 insertion cases.

\begin{table}[t]
\caption{User study results. Percentage of participants favoring each method for realism, consistency, and fidelity. }
\centering
\resizebox{\linewidth}{!}{
\begin{tabular}{l c c c}
    Method & Realism $\uparrow$ & Consistency $\uparrow$ & Fidelity $\uparrow$ \\
    \hline
    AnyV2V~\cite{ku2024anyv2v} & 8.7\% & 8.7\% & 6.7\% \\
    ReVideo~\cite{mou-24} & 4.0\% & 3.6\% &  4.4\%\\
    ConsistI2V~\cite{ren2024consisti2v} & 5.6\% & 3.6\% & 4.8\% \\
    \hline
    Ours & \textbf{81.7\%} & \textbf{84.1\%} & \textbf{84.1\%} \\
\end{tabular}
}
\label{tab:user_study}
\end{table}
\begin{table}[t]
\caption{Quantitative results. DeQA Score~\cite{deqa_score}, Temporal Consistency~\cite{esser2023structure}, and CLIP Score are used to evaluate visual quality, temporal coherence, and the fidelity to the bracelet, respectively.}
\centering
\resizebox{0.9\linewidth}{!}{
\begin{tabular}{l c c c}
    Method & \makecell{DeQA\\Score} $\uparrow$ &  \makecell{Temporal \\ Consistency} $\uparrow$ & \makecell{CLIP\\Score} $\uparrow$\\
    \hline
    AnyV2V~\cite{ku2024anyv2v} & 3.41 & 0.971 & 0.659 \\
    ReVideo~\cite{mou-24} & 3.17 & 0.977 & 0.661 \\
    ConsistI2V~\cite{ren2024consisti2v} & 2.44 & 0.976 & 0.638 \\
    \hline
    Ours & \textbf{3.66} & \textbf{0.984} & \textbf{0.662} \\
\end{tabular}
}
\label{tab:quant_comp}
\end{table}
\noindent\textbf{User Study. }
We begin our evaluation with a user study to compare our method against three state-of-the-art video editing and generation approaches: AnyV2V~\cite{ku2024anyv2v}, ReVideo~\cite{mou-24}, and ConsistI2V~\cite{ren2024consisti2v}. We conduct a user study comparing our method with three state-of-the-art approaches: \emph{AnyV2V} and \emph{ReVideo} for video editing, and \emph{ConsistI2V} for image-to-video generation. We use \emph{Anydoor}~\cite{chen2024anydoor} to insert the bracelet into the first frame for all methods. To ensure a fair comparison, we fine-tune Anydoor on our collected bracelet training data. For AnyV2V and ReVideo, we propagate the edit from the first frame to subsequent frames using their frameworks. For ConsistI2V, we generate the remaining frames based on the first frame edited by Anydoor.

We recruit 36 participants and provide them with a randomized set of 7 video groups. For each insertion case, we present results from three baseline methods and our method. Each participant is instructed to evaluate the videos by selecting the best option in each group based on three key aspects: ``Which video makes the inserted object look most like it belongs in the scene, with realistic lighting'' (Realism), ``Which video shows the object moving most smoothly and naturally, without flickering or sudden changes'' (Temporal Consistency), and ``Which video best preserves the original appearance of the object, such as texture and shape'' (Fidelity). The user study results, as shown in Table~\ref{tab:user_study}, demonstrate that our method is significantly preferred over the competing approaches.

\noindent\textbf{Qualitative Comparisons. } 
Fig.~\ref{fig:visual_comparison} provides visual comparisons between our method and the baseline methods: AnyV2V, ReVideo, and ConsistI2V. As shown in the figure, the baselines suffer from lighting inconsistencies due to their heavy reliance on the first-frame editing capabilities of \emph{Anydoor}, as well as distortions in both the bracelet and wrist during motion. In contrast, our method leverages 3D Gaussian rendering to ensure temporal consistency, preserving the natural appearance of the wrist and maintaining realistic lighting interactions for the bracelet. 

\noindent\textbf{Quantitative Comparisons. } We employ three metrics for quantitative evaluation: 1) DeQA Score~\cite{deqa_score}, a state-of-the-art image quality assessment method; 2) Temporal Consistency~\cite{esser2023structure}, calculated by analyzing the cosine similarity of CLIP~\cite{radford2021learning} embeddings between consecutive frames to evaluate motion smoothness; and 3) CLIP Score, which measures the semantic alignment between generated frames and the reference bracelet image through CLIP~\cite{radford2021learning} embedding similarity. As shown in Tab.~\ref{tab:quant_comp}, our approach achieves superior performance across all evaluation metrics, particularly excelling in visual quality and temporal coherence.

\begin{figure}[t]
    \centering
    \includegraphics[width=\linewidth]{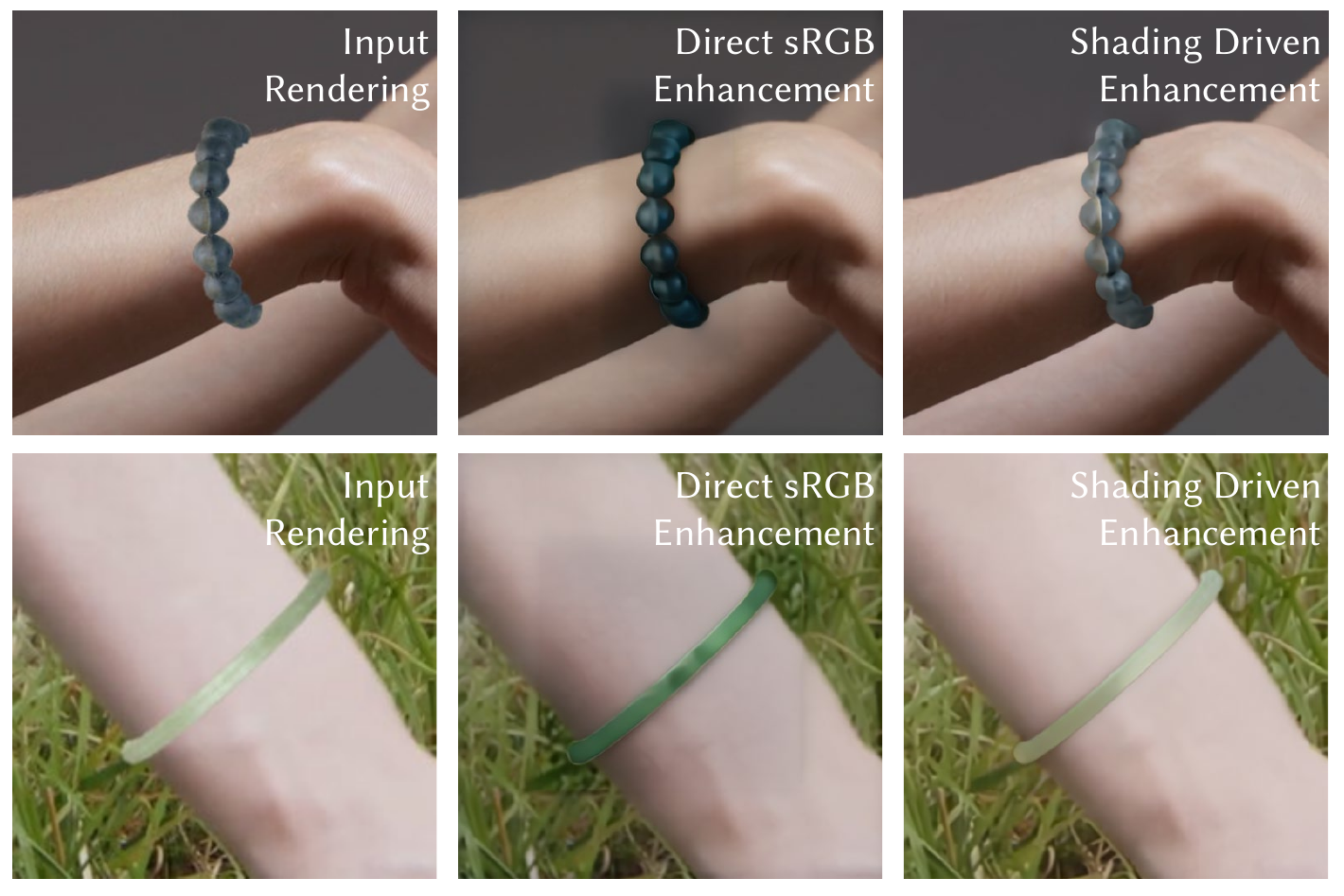}
    \caption{Ablation Study on Shading-Driven Enhancement. We compare our shading-driven pipeline with a baseline that directly predicts sRGB enhancement. Our approach produces more realistic and detailed results, demonstrating the importance of explicit shading refinement for seamless object insertion.}
    \label{fig:ablation_results_shading}
\end{figure}
\begin{figure}[ht]
    \centering
    \includegraphics[width=\linewidth]{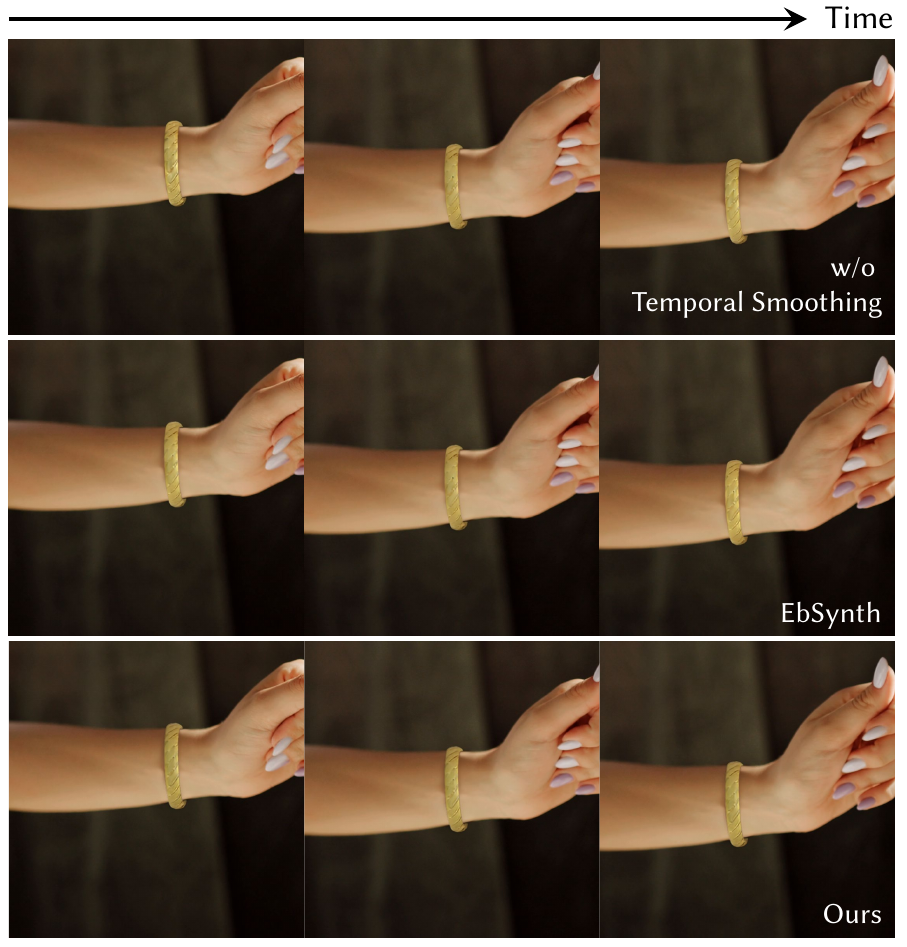}
    \caption{Ablation Study on Temporal Smoothing. We compare our approach with a version without temporal smoothing and a baseline using \emph{EbSynth}. Our 3DGS-based smoothing achieves smooth transitions while preserving fine details. Please refer to our supplementary video for a better comparison.}
    \label{fig:ablation_results_smooth}
\end{figure}
\noindent\textbf{Ablation for Shading-Driven Enhancement. }
To evaluate the effectiveness of our shading-driven realism enhancement pipeline, we compare it with a baseline that directly predicts the sRGB enhancement of the bracelet and shadow. As shown in Fig.~\ref{fig:ablation_results_shading}, the baseline produces bracelets that appear flat and poorly integrated into the scene, with inconsistent lighting and a lack of fine details. In contrast, our shading-driven pipeline, which explicitly enhances shading and refines the results in sRGB space, generates bracelets that seamlessly blend into the scene, with realistic lighting interactions and detailed textures.

\noindent\textbf{Ablation for Temporal Consistency. }
To validate the effectiveness of our temporal smoothing strategy, we compare our full approach with two alternatives: (1) a version without temporal smoothing and (2) a baseline that uses \emph{EbSynth}~\cite{Jamriska2018} for smoothing. As shown in Fig.~\ref{fig:ablation_results_smooth}, the version without temporal smoothing exhibits noticeable flickering and inconsistent motion, while \emph{EbSynth}~\cite{Jamriska2018} produces smoother transitions but struggles to preserve fine details. In contrast, our 3DGS-based smoothing achieves both smooth transitions and high fidelity, maintaining the bracelet's appearance and adapting naturally to the scene.

\section{Conclusion and Limitations}
In this work, we address the challenging problem of inserting 3D objects into videos while maintaining both temporal consistency and realistic lighting. By combining the strengths of 3D rendering techniques and 2D diffusion models, we propose a novel pipeline that leverages 3D Gaussian Splatting (3DGS) for temporal coherence and a diffusion-based enhancement model for photorealistic refinement. We demonstrate realistic insertion results in the bracelet-wearing scenario. The main limitation of our method is its difficulty in handling poor-quality background videos, such as those with motion blur or severe artifacts. This limitation could potentially be addressed by integrating video restoration models. In the future, we plan to extend our framework to general object insertion and further enhance its adaptability and robustness in complex scenarios.

\section*{Acknowledgment}
This research is supported by RGC Early Career Scheme (ECS) No. 24209224. We would like to thank Yaokun Li for his assistance with the user study and Yumeng Shi for her valuable suggestions on the figures and demo. We also thank the reviewers for their insightful comments.

{
    \small
    \bibliographystyle{ieeenat_fullname}
    \bibliography{main}

\begin{thebibliography}{43}
\providecommand{\natexlab}[1]{#1}
\providecommand{\url}[1]{\texttt{#1}}
\expandafter\ifx\csname urlstyle\endcsname\relax
  \providecommand{\doi}[1]{doi: #1}\else
  \providecommand{\doi}{doi: \begingroup \urlstyle{rm}\Url}\fi

\bibitem[Bao et~al.(2022)Bao, Long, Fu, Liu, Li, Wu, and Xiao]{bao2022deep}
Zhongyun Bao, Chengjiang Long, Gang Fu, Daquan Liu, Yuanzhen Li, Jiaming Wu, and Chunxia Xiao.
\newblock Deep image-based illumination harmonization.
\newblock In \emph{Proceedings of the IEEE/CVF Conference on Computer Vision and Pattern Recognition}, pages 18542--18551, 2022.

\bibitem[Bhattad and Forsyth(2022)]{bhattad2022cut}
Anand Bhattad and David~A Forsyth.
\newblock Cut-and-paste object insertion by enabling deep image prior for reshading.
\newblock In \emph{2022 International Conference on 3D Vision (3DV)}, pages 332--341. IEEE, 2022.

\bibitem[Careaga and Aksoy(2024)]{careaga2024colorful}
Chris Careaga and Ya{\u{g}}{\i}z Aksoy.
\newblock Colorful diffuse intrinsic image decomposition in the wild.
\newblock \emph{ACM Transactions on Graphics (TOG)}, 43\penalty0 (6):\penalty0 1--12, 2024.

\bibitem[Careaga et~al.(2023)Careaga, Miangoleh, and Aksoy]{careagaCompositing}
Chris Careaga, S.~Mahdi~H. Miangoleh, and Ya\u{g}{\i}z Aksoy.
\newblock Intrinsic harmonization for illumination-aware compositing.
\newblock In \emph{Proc. SIGGRAPH Asia}, 2023.

\bibitem[Chen et~al.(2024)Chen, Huang, Liu, Shen, Zhao, and Zhao]{chen2024anydoor}
Xi Chen, Lianghua Huang, Yu Liu, Yujun Shen, Deli Zhao, and Hengshuang Zhao.
\newblock Anydoor: Zero-shot object-level image customization.
\newblock In \emph{Proceedings of the IEEE/CVF Conference on Computer Vision and Pattern Recognition (CVPR)}, 2024.

\bibitem[Cong et~al.(2022)Cong, Tao, Niu, Liang, Gao, Sun, and Zhang]{cong2022high}
Wenyan Cong, Xinhao Tao, Li Niu, Jing Liang, Xuesong Gao, Qihao Sun, and Liqing Zhang.
\newblock High-resolution image harmonization via collaborative dual transformations.
\newblock In \emph{Proceedings of the IEEE/CVF Conference on Computer Vision and Pattern Recognition}, pages 18470--18479, 2022.

\bibitem[Doersch(2016)]{doersch2016tutorial}
Carl Doersch.
\newblock Tutorial on variational autoencoders.
\newblock \emph{arXiv preprint arXiv:1606.05908}, 2016.

\bibitem[Esser et~al.(2023)Esser, Chiu, Atighehchian, Granskog, and Germanidis]{esser2023structure}
Patrick Esser, Johnathan Chiu, Parmida Atighehchian, Jonathan Granskog, and Anastasis Germanidis.
\newblock Structure and content-guided video synthesis with diffusion models.
\newblock In \emph{Proceedings of the IEEE/CVF international conference on computer vision}, pages 7346--7356, 2023.

\bibitem[Fischler and Bolles(1981)]{fischler1981random}
Martin~A Fischler and Robert~C Bolles.
\newblock Random sample consensus: a paradigm for model fitting with applications to image analysis and automated cartography.
\newblock \emph{Communications of the ACM}, 24\penalty0 (6):\penalty0 381--395, 1981.

\bibitem[Gao et~al.(2025)Gao, Ding, Cai, Huang, Wang, and Xue]{gao2025lora}
Chenjian Gao, Lihe Ding, Xin Cai, Zhanpeng Huang, Zibin Wang, and Tianfan Xue.
\newblock Lora-edit: Controllable first-frame-guided video editing via mask-aware lora fine-tuning.
\newblock \emph{arXiv preprint arXiv:2506.10082}, 2025.

\bibitem[Guo et~al.(2021)Guo, Zheng, Jiang, Gu, and Zheng]{guo2021intrinsic}
Zonghui Guo, Haiyong Zheng, Yufeng Jiang, Zhaorui Gu, and Bing Zheng.
\newblock Intrinsic image harmonization.
\newblock In \emph{Proceedings of the ieee/cvf conference on computer vision and pattern recognition}, pages 16367--16376, 2021.

\bibitem[Guo et~al.(2022)Guo, Gu, Zheng, Dong, and Zheng]{guo2022transformer}
Zonghui Guo, Zhaorui Gu, Bing Zheng, Junyu Dong, and Haiyong Zheng.
\newblock Transformer for image harmonization and beyond.
\newblock \emph{IEEE transactions on pattern analysis and machine intelligence}, 45\penalty0 (11):\penalty0 12960--12977, 2022.

\bibitem[Hu et~al.(2021)Hu, Nsampi, Wang, and Wang]{hu2021neursf}
Zhongyun Hu, Ntumba~Elie Nsampi, Xue Wang, and Qing Wang.
\newblock Neursf: Neural shading field for image harmonization.
\newblock \emph{arXiv preprint arXiv:2112.01314}, 2021.

\bibitem[Jamriska(2018)]{Jamriska2018}
Ondrej Jamriska.
\newblock Ebsynth: Fast example-based image synthesis and style transfer.
\newblock \url{https://github.com/jamriska/ebsynth}, 2018.

\bibitem[Karaev et~al.(2024)Karaev, Makarov, Wang, Neverova, Vedaldi, and Rupprecht]{karaev24cotracker3}
Nikita Karaev, Iurii Makarov, Jianyuan Wang, Natalia Neverova, Andrea Vedaldi, and Christian Rupprecht.
\newblock Cotracker3: Simpler and better point tracking by pseudo-labelling real videos.
\newblock In \emph{Proc. {arXiv:2410.11831}}, 2024.

\bibitem[Ke et~al.(2022)Ke, Sun, Zhu, Xu, and Lau]{ke2022harmonizer}
Zhanghan Ke, Chunyi Sun, Lei Zhu, Ke Xu, and Rynson~WH Lau.
\newblock Harmonizer: Learning to perform white-box image and video harmonization.
\newblock In \emph{European Conference on Computer Vision}, pages 690--706. Springer, 2022.

\bibitem[Kerbl et~al.(2023)Kerbl, Kopanas, Leimk{\"u}hler, and Drettakis]{kerbl3Dgaussians}
Bernhard Kerbl, Georgios Kopanas, Thomas Leimk{\"u}hler, and George Drettakis.
\newblock 3d gaussian splatting for real-time radiance field rendering.
\newblock \emph{ACM Transactions on Graphics}, 42\penalty0 (4), 2023.

\bibitem[Kholgade et~al.(2014)Kholgade, Simon, Efros, and Sheikh]{Natasha-14}
Natasha Kholgade, Tomas Simon, Alexei Efros, and Yaser Sheikh.
\newblock 3d object manipulation in a single photograph using stock 3d models.
\newblock \emph{ACM Transactions on graphics (TOG)}, 33\penalty0 (4):\penalty0 1--12, 2014.

\bibitem[Kinga et~al.(2015)Kinga, Adam, et~al.]{kinga2015method}
Diederik Kinga, Jimmy~Ba Adam, et~al.
\newblock A method for stochastic optimization.
\newblock In \emph{International conference on learning representations (ICLR)}. San Diego, California;, 2015.

\bibitem[Ku et~al.(2024)Ku, Wei, Ren, Yang, and Chen]{ku2024anyv2v}
Max Ku, Cong Wei, Weiming Ren, Huan Yang, and Wenhu Chen.
\newblock Anyv2v: A plug-and-play framework for any video-to-video editing tasks.
\newblock \emph{arXiv preprint arXiv:2403.14468}, 2024.

\bibitem[Li and Snavely(2018)]{li2018megadepth}
Zhengqi Li and Noah Snavely.
\newblock Megadepth: Learning single-view depth prediction from internet photos.
\newblock In \emph{Proceedings of the IEEE conference on computer vision and pattern recognition}, pages 2041--2050, 2018.

\bibitem[Liao et~al.(2019)Liao, Karsch, Zhang, and Forsyth]{liao2019approximate}
Zicheng Liao, Kevin Karsch, Hongyi Zhang, and David Forsyth.
\newblock An approximate shading model with detail decomposition for object relighting.
\newblock \emph{International Journal of Computer Vision}, 127:\penalty0 22--37, 2019.

\bibitem[Loshchilov(2017)]{loshchilov2017decoupled}
I Loshchilov.
\newblock Decoupled weight decay regularization.
\newblock \emph{arXiv preprint arXiv:1711.05101}, 2017.

\bibitem[Martin~Garcia et~al.(2025)Martin~Garcia, Abou~Zeid, Schmidt, de~Geus, Hermans, and Leibe]{martingarcia2024diffusione2eft}
Gonzalo Martin~Garcia, Karim Abou~Zeid, Christian Schmidt, Daan de Geus, Alexander Hermans, and Bastian Leibe.
\newblock Fine-tuning image-conditional diffusion models is easier than you think.
\newblock In \emph{Proceedings of the IEEE/CVF Winter Conference on Applications of Computer Vision (WACV)}, 2025.

\bibitem[Mou et~al.(2024)Mou, Cao, Wang, Zhang, Shan, and Zhang]{mou-24}
Chong Mou, Mingdeng Cao, Xintao Wang, Zhaoyang Zhang, Ying Shan, and Jian Zhang.
\newblock Revideo: Remake a video with motion and content control.
\newblock \emph{NeurIPS}, 2024.

\bibitem[Ouyang et~al.(2024)Ouyang, Dong, Yang, Si, and Pan]{ouyang-24}
Wenqi Ouyang, Yi Dong, Lei Yang, Jianlou Si, and Xingang Pan.
\newblock I2vedit: First-frame-guided video editing via image-to-video diffusion models.
\newblock In \emph{SIGGRAPH Asia 2024 Conference Papers}. Association for Computing Machinery, 2024.

\bibitem[Piccinelli et~al.(2024)Piccinelli, Yang, Sakaridis, Segu, Li, Van~Gool, and Yu]{piccinelli2024unidepth}
Luigi Piccinelli, Yung-Hsu Yang, Christos Sakaridis, Mattia Segu, Siyuan Li, Luc Van~Gool, and Fisher Yu.
\newblock {U}ni{D}epth: Universal monocular metric depth estimation.
\newblock In \emph{Proceedings of the IEEE/CVF Conference on Computer Vision and Pattern Recognition (CVPR)}, 2024.

\bibitem[Radford et~al.(2021)Radford, Kim, Hallacy, Ramesh, Goh, Agarwal, Sastry, Askell, Mishkin, Clark, et~al.]{radford2021learning}
Alec Radford, Jong~Wook Kim, Chris Hallacy, Aditya Ramesh, Gabriel Goh, Sandhini Agarwal, Girish Sastry, Amanda Askell, Pamela Mishkin, Jack Clark, et~al.
\newblock Learning transferable visual models from natural language supervision.
\newblock In \emph{International conference on machine learning}, pages 8748--8763. PmLR, 2021.

\bibitem[Rav-Acha et~al.(2008)Rav-Acha, Kohli, Rother, and Fitzgibbon]{Alex-07}
Alex Rav-Acha, Pushmeet Kohli, Carsten Rother, and Andrew Fitzgibbon.
\newblock Unwrap mosaics: a new representation for video editing.
\newblock \emph{ACM Trans. Graph.}, 27\penalty0 (3):\penalty0 1–11, 2008.

\bibitem[Ren et~al.(2024)Ren, Yang, Zhang, Wei, Du, Huang, and Chen]{ren2024consisti2v}
Weiming Ren, Huan Yang, Ge Zhang, Cong Wei, Xinrun Du, Wenhao Huang, and Wenhu Chen.
\newblock Consisti2v: Enhancing visual consistency for image-to-video generation.
\newblock \emph{arXiv preprint arXiv:2402.04324}, 2024.

\bibitem[Rombach et~al.(2022)Rombach, Blattmann, Lorenz, Esser, and Ommer]{Rombach_2022_CVPR}
Robin Rombach, Andreas Blattmann, Dominik Lorenz, Patrick Esser, and Bj\"orn Ommer.
\newblock High-resolution image synthesis with latent diffusion models.
\newblock In \emph{Proceedings of the IEEE/CVF Conference on Computer Vision and Pattern Recognition (CVPR)}, pages 10684--10695, 2022.

\bibitem[Ronneberger et~al.(2015)Ronneberger, Fischer, and Brox]{ronneberger2015u}
Olaf Ronneberger, Philipp Fischer, and Thomas Brox.
\newblock U-net: Convolutional networks for biomedical image segmentation.
\newblock In \emph{Proceedings of Medical Image Computing and Computer-Assisted Intervention (MICCAI)}, 2015.

\bibitem[Song et~al.(2023)Song, Zhang, Lin, Cohen, Price, Zhang, Kim, and Aliaga]{song2023objectstitch}
Yizhi Song, Zhifei Zhang, Zhe Lin, Scott Cohen, Brian Price, Jianming Zhang, Soo~Ye Kim, and Daniel Aliaga.
\newblock Objectstitch: Object compositing with diffusion model.
\newblock In \emph{Proceedings of the IEEE/CVF Conference on Computer Vision and Pattern Recognition}, pages 18310--18319, 2023.

\bibitem[Song et~al.(2024)Song, Zhang, Lin, Cohen, Price, Zhang, Kim, Zhang, Xiong, and Aliaga]{song2024imprint}
Yizhi Song, Zhifei Zhang, Zhe Lin, Scott Cohen, Brian Price, Jianming Zhang, Soo~Ye Kim, He Zhang, Wei Xiong, and Daniel Aliaga.
\newblock Imprint: Generative object compositing by learning identity-preserving representation.
\newblock In \emph{Proceedings of the IEEE/CVF Conference on Computer Vision and Pattern Recognition}, pages 8048--8058, 2024.

\bibitem[Tu et~al.(2025)Tu, Luo, Chen, Ji, Bai, and Zhao]{tu-25}
Yuanpeng Tu, Hao Luo, Xi Chen, Sihui Ji, Xiang Bai, and Hengshuang Zhao.
\newblock Videoanydoor: High-fidelity video object insertion with precise motion control, 2025.

\bibitem[van~den Hengel et~al.(2007)van~den Hengel, Dick, Thorm\"{a}hlen, Ward, and Torr]{Anton-07}
Anton van~den Hengel, Anthony Dick, Thorsten Thorm\"{a}hlen, Ben Ward, and Philip H.~S. Torr.
\newblock Videotrace: rapid interactive scene modelling from video.
\newblock \emph{ACM Trans. Graph.}, 26\penalty0 (3):\penalty0 86–es, 2007.

\bibitem[Wang et~al.(2023)Wang, Gharbi, Zhang, Xia, and Shechtman]{wang2023semi}
Ke Wang, Micha{\"e}l Gharbi, He Zhang, Zhihao Xia, and Eli Shechtman.
\newblock Semi-supervised parametric real-world image harmonization.
\newblock In \emph{Proceedings of the IEEE/CVF Conference on Computer Vision and Pattern Recognition}, pages 5927--5936, 2023.

\bibitem[Weng et~al.(2024)Weng, Zhang, Li, Li, Shi, et~al.]{weng2024cad}
Shuchen Weng, Peixuan Zhang, Yu Li, Si Li, Boxin Shi, et~al.
\newblock L-cad: Language-based colorization with any-level descriptions using diffusion priors.
\newblock \emph{Advances in Neural Information Processing Systems}, 36, 2024.

\bibitem[Winter et~al.(2024)Winter, Cohen, Fruchter, Pritch, Rav-Acha, and Hoshen]{Winter-24}
Daniel Winter, Matan Cohen, Shlomi Fruchter, Yael Pritch, Alex Rav-Acha, and Yedid Hoshen.
\newblock Objectdrop: Bootstrapping counterfactuals for photorealistic object removal and insertion.
\newblock In \emph{ECCV}, 2024.

\bibitem[Yang et~al.(2023{\natexlab{a}})Yang, Gu, Zhang, Zhang, Chen, Sun, Chen, and Wen]{yang-23}
Binxin Yang, Shuyang Gu, Bo Zhang, Ting Zhang, Xuejin Chen, Xiaoyan Sun, Dong Chen, and Fang Wen.
\newblock Paint by example: Exemplar-based image editing with diffusion models.
\newblock In \emph{2023 IEEE/CVF Conference on Computer Vision and Pattern Recognition (CVPR)}, pages 18381--18391, 2023{\natexlab{a}}.

\bibitem[Yang et~al.(2023{\natexlab{b}})Yang, Zhou, Liu, and Loy]{yang2023rerender}
Shuai Yang, Yifan Zhou, Ziwei Liu, and Chen~Change Loy.
\newblock Rerender a video: Zero-shot text-guided video-to-video translation.
\newblock In \emph{SIGGRAPH Asia 2023 Conference Papers}, pages 1--11, 2023{\natexlab{b}}.

\bibitem[You et~al.(2025)You, Cai, Gu, Xue, and Dong]{deqa_score}
Zhiyuan You, Xin Cai, Jinjin Gu, Tianfan Xue, and Chao Dong.
\newblock Teaching large language models to regress accurate image quality scores using score distribution.
\newblock In \emph{IEEE Conference on Computer Vision and Pattern Recognition}, 2025.

\bibitem[Zhang et~al.(2023)Zhang, Duan, Lan, Hong, Zhu, Wang, and Niu]{zhang2023controlcom}
Bo Zhang, Yuxuan Duan, Jun Lan, Yan Hong, Huijia Zhu, Weiqiang Wang, and Li Niu.
\newblock Controlcom: Controllable image composition using diffusion model.
\newblock \emph{arXiv preprint arXiv:2308.10040}, 2023.

\end{thebibliography}
}

\clearpage
\setcounter{page}{1}
\maketitlesupplementary

We highly recommend that readers view the videos on our \textbf{project page} to explore additional results and visualizations. Below, we include technical details omitted from the main text.
\section{Bracelet 3D Motion Calculation}
\label{sec:motion}
Given a 3D Gaussian Splatting (3DGS) model \( \mathcal{G} \) of the bracelet and its initial pose \( \mathbf{P}_1 = (\mathbf{R}_1, \mathbf{T}_1) \) in the first frame, we compute the bracelet’s pose in subsequent frames to align it with the wrist’s motion. We rely on 2D keypoint tracking result of skin near the bracelet, and use them to calculate 3D bracelet motion.

Specifically, this is modeled as a \emph{Perspective-n-Point (PnP)}~\cite{fischler1981random} problem. The input is a set of 2D points \( \{\mathbf{x}_t^i\}_{i=1}^N \) on the skin, initialized in the first frame as \( \{\mathbf{x}_1^i\}_{i=1}^N \), and tracked across frames using \emph{CoTracker}~\cite{karaev24cotracker3}. Then, we lift these 2D points \( \{\mathbf{x}_1^i\} \) to 3D using the depth map \( \mathbf{D}_1 \), camera intrinsics \( \mathbf{K} \), and the initial pose \( \mathbf{P}_1 \):
\[
\mathbf{X}_1^i = \mathbf{P}_1^{-1} \cdot \left( \mathbf{K}^{-1} \begin{bmatrix} \mathbf{x}_1^i \\ \mathbf{D}_1(\mathbf{x}_1^i) \end{bmatrix} \right),
\]
where \( \mathbf{D}_1 \) and \(\mathbf{K}\) are estimated by \emph{UniDepth}~\cite{piccinelli2024unidepth}.
For each frame \( t \), the corresponding pose is computed by solving PnP equation:
\[
\arg\min_{\mathbf{P}_t} \sum_{i=1}^N \|\mathbf{K} \mathbf{P}_t \mathbf{X}_1^i - \mathbf{x}_t^i\|^2.
\]
Finally, we smoothed the poses by applying a bilateral filter to the translation vector \( \mathbf{T} \) and quaternion representations of the rotation \( \mathbf{R} \) to get the final \( {\mathbf{P}_t} \).

\section{Occlusion Handling}
\label{sec:occlusion}
To ensure a correct depth ordering between human, background, and bracelet, we use monocular depth maps \( \mathbf{D}_t \) from \emph{UniDepth} as the 3D context to handle the occlusion. To align depth across frames, we compute a scale factor \( s_t \) for each frame \( t \) using the calculated pose \( \mathbf{P}_t \):
\[
\arg\min_{s_t} \sum_{i=1}^N \|s_t \cdot \mathbf{D}_t(\mathbf{x}_t^i) - [\mathbf{P}_t \mathbf{X}_1^i]_z\|^2.
\]
The solution to this optimization problem is:
\[
s_t = \frac{\sum_{i=1}^N \mathbf{D}_t(\mathbf{x}_t^i) \cdot [\mathbf{P}_t \mathbf{X}_1^i]_z}{\sum_{i=1}^N \mathbf{D}_t(\mathbf{x}_t^i)^2}.
\]

Once the depth maps are aligned, we compute the occlusion mask \( \mathbf{M}_t \) for each frame by comparing the scene depth \( \mathbf{D}_t \) with the bracelet’s depth rendering \(\mathbf{D}_t^\text{bracelet}\). 
\[
\mathbf{O}_t(\mathbf{x}) = \begin{cases} 
1 & \text{if } s_t \cdot \mathbf{D}_t(\mathbf{x}) < \mathbf{D}_t^\text{bracelet}(\mathbf{x}), \\
0 & \text{otherwise}.
\end{cases}
\]
To avoid hard edges and aliasing artifacts, we further apply a Gaussian blur to the initial occlusion map \( \mathbf{O}_t \), yielding the soft mask \( \mathbf{M}_t = G_\sigma * \mathbf{O}_t \), where \( G_\sigma \) is a Gaussian kernel with standard deviation \( \sigma \).

Finally we compute a preview of the bracelet in the current frame:
\[
\mathbf{I}_t^\text{preview}(\mathbf{x}) = \mathbf{M}_t(\mathbf{x}) \cdot \mathbf{I}_t^\text{bracelet}(\mathbf{x}) + (1 - \mathbf{M}_t(\mathbf{x})) \cdot \mathbf{I}_t^\text{scene}(\mathbf{x}),
\]
where \( \mathbf{I}_t^\text{bracelet} \) is the rendered image of the bracelet at pose \( \mathbf{P}_t \), and \( \mathbf{I}_t^\text{scene} \) is the original scene image. This preview is an intermediate output, with further refinements applied later to enhance realism.

\section{Optimizing 3D Gaussian for Smoothing}
To achieve smooth temporal transitions while preserving the geometric structure of the bracelet, we optimize only the spherical harmonics (SH) coefficients associated with each splat in the 3D Gaussian Splatting (3DGS) model~\cite{kerbl3Dgaussians}. The SH coefficients encode view-dependent color information for each Gaussian splat, allowing us to adjust the appearance of the bracelet without modifying its geometric attributes (position, scale, rotation, and opacity). For each splat \( i \), let \( \mathbf{c}_i(\mathbf{d}) \) represent its view-dependent color, which is a function of the viewing direction \( \mathbf{d} \). This color is computed using the spherical harmonics basis functions \( Y_l^m(\mathbf{d}) \) and the corresponding SH coefficients \( \mathbf{s}_i \), where \( s_{i,l}^m \) are the SH coefficients for splat \( i \), with \( l \) and \( m \) representing the degree and order of the spherical harmonics, respectively, \( Y_l^m(\mathbf{d}) \) are the spherical harmonics basis functions evaluated at the viewing direction \( \mathbf{d} \). 

During optimization, we update the SH coefficients \( \mathbf{s}_i \) for each splat \( i \) to minimize the photometric error between the rendered image and the refined reference image \( \mathbf{I}_t^\text{refined} \). This ensures that the color and shading of the bracelet are smoothly adjusted while keeping the geometric structure fixed. The optimization can be expressed as:
\[
\mathbf{s}_i^* = \operatorname*{argmin}_{\mathbf{s}_i} \sum_{k=t-W/2}^{t+W/2} w(k - t) \cdot \| \mathcal{R}(\mathbf{K}, \mathbf{P}_k, \mathcal{G}(\mathbf{s}_i)) - \mathbf{I}_t^\text{refined} \|^2,
\]
where \( \mathcal{G}(\mathbf{s}_i) \) denotes the 3DGS model with updated SH coefficients \( \mathbf{s}_i \) for each splat. We employ the Adam~\cite{kinga2015method} optimizer to solve this optimization problem, with a learning rate of \( 10^{-2} \) for the DC (direct current) component (\( l = 0 \)) and \( 10^{-4} \) for the AC (alternating current) components (\( l > 0 \)) of the spherical harmonics.

\end{document}